\documentclass[final,3p,11pt]{elsarticle}

\usepackage{graphicx}
\usepackage{mathptmx} 
\usepackage{latexsym}

\usepackage{amssymb}
\usepackage{lineno}
\usepackage{subfig}
\usepackage{algorithm}
\usepackage[noend]{algpseudocode}
\usepackage{multirow}
\usepackage{booktabs}
\usepackage{multicol}
\usepackage{array}
\usepackage{bm}
\usepackage{amsmath}

\usepackage{hyperref}

\usepackage{natbib}
\biboptions{authoryear}

\usepackage{tikz}
\usepackage{pgfplots}
\usetikzlibrary{patterns}
\usepackage{csvsimple}
\usepackage{pgfplotstable}

\usepackage[markup=none,commentmarkup=footnote]{changes}
\usepackage{color}
\definechangesauthor[color=magenta]{AE}

\DeclareMathOperator{\sign}{sgn}

\definecolor{blue}{RGB}{8,141,165}
\definecolor{olivegreen}{RGB}{85,107,47}
\definecolor{orange}{RGB}{229,148,0}
\definecolor{marine}{RGB}{0,32,96}
\definecolor{maroon}{RGB}{178, 50, 50}

\pgfplotsset{every tick label/.append style={font=\tiny}}
\pgfplotsset{
	box plot width/.initial=4em,
	box plot/.style={
		/pgfplots/.cd,
		black,
		only marks,
		mark=-,
		mark size=\pgfkeysvalueof{/pgfplots/box plot width},
		/pgfplots/error bars/.cd,
		y dir=plus,
		y explicit,
	},
	box plot box/.style={
		/pgfplots/error bars/draw error bar/.code 2 args={%
			\draw [line width=0.20mm]  ##1 -- ++(\pgfkeysvalueof{/pgfplots/box plot width},0pt) |- ##2 -- ++(-\pgfkeysvalueof{/pgfplots/box plot width},0pt) |- ##1 -- cycle;
		},
		/pgfplots/table/.cd,
		y index=2,
		y error expr={\thisrowno{3}-\thisrowno{2}},
		/pgfplots/box plot
	},
	box plot top whisker/.style={
		/pgfplots/error bars/draw error bar/.code 2 args={%
			\pgfkeysgetvalue{/pgfplots/error bars/error mark}%
			{\pgfplotserrorbarsmark}%
			\pgfkeysgetvalue{/pgfplots/error bars/error mark options}%
			{\pgfplotserrorbarsmarkopts}%
			\path ##1 -- ##2;
		},
		/pgfplots/table/.cd,
		y index=4,
		y error expr={\thisrowno{2}-\thisrowno{4}},
		/pgfplots/box plot
	},
	box plot bottom whisker/.style={
		/pgfplots/error bars/draw error bar/.code 2 args={%
			\pgfkeysgetvalue{/pgfplots/error bars/error mark}%
			{\pgfplotserrorbarsmark}%
			\pgfkeysgetvalue{/pgfplots/error bars/error mark options}%
			{\pgfplotserrorbarsmarkopts}%
			\path ##1 -- ##2;
		},
		/pgfplots/table/.cd,
		y index=5,
		y error expr={\thisrowno{3}-\thisrowno{5}},
		/pgfplots/box plot
	},
	box plot median/.style={
		/pgfplots/box plot
	}
}
\pgfplotsset{yticklabel style={text width=1.0em,align=right}}
\pgfplotsset{compat=1.12}

\pgfplotstableread[col sep = comma]{plot_data/tsp_varying_size_gap.csv}\TSPSizeGap
\pgfplotstableread[col sep = comma]{plot_data/tsp_varying_size_p.csv}\TSPSizeP

\pgfplotstableread[col sep = comma]{plot_data/tsp_10_gap.csv}\TSPTenGap
\pgfplotstableread[col sep = comma]{plot_data/tsp_10_p.csv}\TSPTenP
\pgfplotstableread[col sep = comma]{plot_data/tsp_100_gap.csv}\TSPHunGap
\pgfplotstableread[col sep = comma]{plot_data/tsp_100_p.csv}\TSPHunP

\pgfplotstableread[col sep = comma]{plot_data/atsp_10_gap.csv}\ATSPTenGap
\pgfplotstableread[col sep = comma]{plot_data/atsp_10_p.csv}\ATSPTenP
\pgfplotstableread[col sep = comma]{plot_data/atsp_100_gap.csv}\ATSPHunGap
\pgfplotstableread[col sep = comma]{plot_data/atsp_100_p.csv}\ATSPHunP

\pgfplotstableread[col sep = comma]{plot_data/sop_10_gap.csv}\SOPTenGap
\pgfplotstableread[col sep = comma]{plot_data/sop_10_p.csv}\SOPTenP
\pgfplotstableread[col sep = comma]{plot_data/sop_100_gap.csv}\SOPHunGap
\pgfplotstableread[col sep = comma]{plot_data/sop_100_p.csv}\SOPHunP

\pgfplotstableread[col sep = comma]{plot_data/variants_mean_gap.csv}\VariantMeanGap
\pgfplotstableread[col sep = comma]{plot_data/variants_mean_p.csv}\VariantMeanP

\pgfplotstableread[col sep = comma]{plot_data/CLKeasy_results.csv}\CLKeasy
\pgfplotstableread[col sep = comma]{plot_data/CLKhard_results.csv}\CLKhard
\pgfplotstableread[col sep = comma]{plot_data/LKCCeasy_results.csv}\LKCCeasy
\pgfplotstableread[col sep = comma]{plot_data/LKCChard_results.csv}\LKCChard
\pgfplotstableread[col sep = comma]{plot_data/random_results.csv}\random

\pgfplotstableread[col sep = comma]{plot_data/tsp_cplex_results_6.csv}\TSPCPLEX

\usepackage{lipsum}
\makeatletter
\def\ps@pprintTitle{%
	\let\@oddhead\@empty
	\let\@evenhead\@empty
	\def\@oddfoot{}%
	\let\@evenfoot\@oddfoot}
\makeatother

\begin{document}
	







\begin{frontmatter}
	
\title{Generalization of Machine Learning for Problem Reduction: A Case Study on Travelling Salesman Problems}
\author[label1]{Yuan Sun}
\ead{yuan.sun@rmit.edu.au}
\author[label2]{Andreas Ernst}
\ead{andreas.ernst@monash.edu}
\author[label1]{Xiaodong Li}
\ead{xiaodong.li@rmit.edu.au}
\author[label1]{Jake Weiner}
\ead{jake.weiner@rmit.edu.au}
\address[label1]{School of Science, RMIT University, Melbourne, 3001, Victoria, Australia}
\address[label2]{School of Mathematical Sciences, Monash University, Clayton, 3800, Victoria, Australia}



\begin{abstract}
	
	Combinatorial optimization plays an important role in real-world problem solving. In the big data era, the dimensionality of a combinatorial optimization problem is usually very large, which poses a significant challenge to existing solution methods. In this paper, we examine the generalization capability of a machine learning model for problem reduction on the classic travelling salesman problems (TSP). We demonstrate that our method can greedily remove decision variables from an optimization problem that are predicted not to be part of an optimal solution. More specifically, we investigate our model's capability to generalize on test instances that have not been seen during the training phase. We consider three scenarios where training and test instances are different in terms of: 1) problem characteristics; 2) problem sizes; and 3) problem types. Our experiments show that this machine learning based technique can generalize reasonably well over a wide range of TSP test instances with different characteristics or sizes. While the accuracy of predicting unused variables naturally deteriorates as a test instance is further away from the training set, we observe that even when tested on a different TSP problem variant, the machine learning model still makes useful predictions about which variables can be eliminated without significantly impacting solution quality.
	
\end{abstract}

	
\end{frontmatter}

\section{Introduction}
In the big data era, we are often confronted with optimization problems with thousands or even millions of decision variables, e.g., social network analysis \citep{balasundaram2011clique,gao2018exact}. The large problem size poses significant challenges to existing solution algorithms, especially to generic Mixed Integer Programming (MIP) solvers such as CPLEX, which typically has difficulty in optimally solving or even finding good solutions for such large-scale optimization problems in a reasonable computational time. Moreover in many practical applications, e.g., trip planning \citep{friggstad2018orienteering}, we need to provide a high-quality solution to users within a few seconds. This is hard to achieve especially when the problem size is very large, which necessitates the use of an effective problem reduction technique that can significantly prune the search space but still capture an optimal (or near-optimal) solution in the reduced space. 

Recently, there has been a growing trend of applying machine learning for problem reduction \citep{li2018combinatorial,lauri2019fine,grassia2019learning,ding2019accelerating,sun2019using}. These machine learning models are typically trained on easy problem instances for which the optimal solution is known, and predict for a given hard unsolved problem instance a subset of decision variables that most likely belong to an optimal solution. By greedily removing decision variables that are not expected to be part of an optimal solution, the original large search space can be significantly reduced to a size that is manageable by an existing solution algorithm. In our recent work \citep{sun2019using}, we have developed such a Machine Learning model for Problem Reduction (MLPR), which builds on statistical measures computed from stochastic sampling of feasible solutions. We have empirically shown that as a prepossessing technique, our MLPR method can significantly improve the performance of existing solution algorithms when used to solve large maximum weight clique problems.

Although the idea of problem reduction using machine learning is generic, it is still unclear whether our MLPR method is also effective on combinatorial problems other than maximum weight clique problems. In this paper, we
examine the effectiveness of our MLPR method on the classic traveling salesman problem (TSP). We consider TSPs on a complete graph where the objective is to search for a shortest route that visits each vertex and returns to the original vertex in the graph. We adapt our MLPR model to predict for each edge whether it belongs to a shortest route, and remove from the complete graph those who do not. The aim is to find a sparse subgraph that still contains a (near) optimal tour. This adaptation is nontrivial, because problem-specific features and sampling methods have to be designed for TSPs. Furthermore we parallelize our MLPR model in this paper, so that the computational time of our MLPR model can be significantly reduced by using  multiple cores.


Since the TSP has been extensively studied and many effective solution algorithms have been developed for solving TSPs \citep{applegate2006concorde,applegate2006traveling,lin1973effective,helsgaun2000effective}, our primal goal here is not to further push the limit of problem solving. Instead, we focus on exploring the generalization capability (more specifically generalization error) of our MLPR model when training and test instances are different. Generalization error is very relevant to real-world problem solving, because in practice the test instances on which a trained model is applied are potentially quite different from the training instances. For example, the routing problem that a navigation company solves on a regular basis might drift over time. 

A significant contribution of this paper is to provide a systematic analysis on the robustness of our MLPR model when such nontrivial changes happen in test instances. We empirically show that our MLPR model generalizes reasonably well to a wide range of test TSP instances with different characteristics or sizes. We also identify where our MLPR model may not perform well, i.e., on the test instances that are deliberately made to be very different from the training instances in terms of problem characteristics. This provides guidance on how to construct a good training set and when to update the training set in practical contexts. 


Taking a step further, we investigate whether the knowledge learned from one variant of TSP instances can be transferred to solving other TSP variants. Our experimental results show that the MLPR model trained on symmetric TSP instances performs fairly well on some of the test instances from other TSP variants, although we do observe a performance degradation when the test TSP variants are gradually moved away from the training TSPs. This indicates it is possible to develop a more generic MLPR model that does not require re-training when applied to different problems (or at least a class of problems).

The remainder of this paper is organized as follows. In Section \ref{Section: Background and Related Work}, we briefly describe the background and methods related to problem reduction. In Section \ref{Section: MLPR}, we adapt our MLPR model to reduce problem size for TSP. Section \ref{Section: Experimental Results} presents the experimental results. The last section concludes the paper and suggests potential research directions for future work. 

\section{Background and Related Work}\label{Section: Background and Related Work}
We briefly describe TSP in Section \ref{Subsection: Travelling Salesman Problem}, and review the problem reduction techniques based on machine learning in Section \ref{Subsection: Problem Reduction Based on Machine Learning}. Because our MLPR model uses support vector machine (SVM)  as the classification algorithm, we will briefly describe SVM in Section \ref{Subsection: Support Vector Machine}. 

\subsection{Travelling Salesman Problem} \label{Subsection: Travelling Salesman Problem}
Given $n$ cities $\{v_1,v_2,\cdots,v_n\}$ and pairwise distance between cities $\{c_{i,j} \, | \, i , j = 1,\cdots, n, \,  i \ne j\}$, the objective of the TSP problem is to find the shortest route that visits each city and returns to the original city. We use $u_i$ to denote the visiting order of city $i$,  and use a binary variable $x_{i,j}$ to denote whether city $j$ is visited directly after city $i$. Without loss of generality, we set $u_1 = 1$ (route starts from city $1$). The Miller-Tucker-Zemlin formulation of TSP can be written as 
\begin{align}
\min_{\bm{x}} & \sum_{i=1}^{n} \sum_{j=1}^{n} c_{i,j} x_{i,j}, \\
s.t. \;\, & \sum_{i=1}^{n} x_{i,j} = 1, \;&& j = 1, 2, \cdots, n; \label{Eq. con1}\\
& \sum_{j=1}^{n} x_{i,j} = 1, \;&& i = 1, 2, \cdots, n; \label{Eq. con2} \\
& u_i - u_j + nx_{i,j} \le n - 1,&& 2 \le i, j \le n; \label{Eq. con3} \\
&  u_i \ge 0, \;&& i =  2, \cdots, n; \\
& x_{i,j} \in \{0, 1\}, \; &&1 \le i, j \le n.
\end{align}
The first two constraints, (\ref{Eq. con1}) and (\ref{Eq. con2}), ensure that each city is arrived at and departed from exactly once; and the constraint (\ref{Eq. con3}) eliminates subtours. More computationally efficient formulations exist, but this is sufficient for logical correctness. Note that of the $n^2$ $x_{ij}$ variables,  exactly $n(n-1)$ must be zero in any feasible solution. Removing such variables that are not part of any optimal solution would give a smaller problem with the same optimum.

The TSP has been intensively studied and many solution algorithms have been developed to solve this problem, e.g., the Concorde exact solver \citep{applegate2006concorde,applegate2006traveling}, the Lin-Kernighan heuristic method \citep{lin1973effective,helsgaun2000effective} and the ``backbone" based heuristics \citep{dong2009effective, jager2014backbone}. Recently, there has been a growing interest in using machine learning to automatically learn a solution algorithm to solve combinatorial optimization problems \citep{bengio2018machine}. The learning-based methods for solving TSP include:  \citet{vinyals2015pointer,bello2016neural,khalil2017learning,deudon2018learning,kool2018attention,wu2019learning,chen2019learning}, to name a few. 

%
%
%
%
%
%


\subsection{Problem Reduction Based on Machine Learning}\label{Subsection: Problem Reduction Based on Machine Learning}

Many combinatorial optimization problems contain a large number of decision variables, most of which are ``irrelevant" to the optimal solution.  For example, in a symmetric TSP with $n$ cities, the total number of binary variables is $n(n-1)/2$, while a shortest route only uses $n$ binary variables (for those the value is $1$). The goal of problem reduction is to identify some of these irrelevant variables and remove them from the original problem, in the hope that the reduced problem can be solved more easily. However, identifying these irrelevant variables is a nontrivial task itself. 

Most of the existing problem reduction methods in mathematical programming are exact approaches, which only remove decision variables that cannot be part of an optimal solution, based on mathematical reasoning and/or computation of an objective bound \citep{jonker1984nonoptimal,hougardy2014edge}. An exact approach guarantees that the reduced problem always contains an original optimal solution, but in many cases it is computationally expensive and/or is not equipped with means to significantly reduce the problem size. 

Fortunately, many combinatorial optimization problems have a ``backbone" structure; that is the optimal solution of a problem is likely to share some components with high-quality solutions \citep{kilby2005backbone,wu2015review}. This makes it possible to statistically quantify which decision variables or solution components are more likely to be part of an optimal solution from sample solutions. This heuristic approach, although it does not have an optimality guarantee, can usually remove a large number of irrelevant decision variables from a given problem instance \citep{fischer2007reducing,sun2019using}. 

Our MLPR method originally proposed in \citep{sun2019using} belongs to the class of heuristic reduction approaches. We use optimally-solved problem instances as training set, and apply machine learning to automatically learn a rule to separate the decision variables that belong to an optimal solution from those who do not (irrelevant variables). We extract computationally-cheap problem features as well as statistical measures computed from random samples of feasible solutions to characterize each decision variable. Based on these features, we predict for each decision variable a likelihood of whether it belongs to an optimal solution. Our MLPR method can be used as a preprocessing technique to remove decision variables that are not expected to be part of an optimal solution from an unseen test problem instance. We will describe our MLPR model in more detail in Section~\ref{Section: MLPR}.  

Closely related to our MLPR method, \citet{lauri2019fine} also developed a machine learning model for problem reduction to list all maximum cliques in a graph. They only use  features directly computed from graph data to characterize a vertex, and remove vertices that are predicted not to be part of a maximum clique. Building on this work, \citet{grassia2019learning} developed a multi-stage pruning technique that can further reduce problem sizes for sparse graphs. They also investigated the effects of removing edges instead of vertices from a graph. The main difference between these methods and ours is that these methods do not use statistical features computed from stochastic samples of feasible solutions. It is worth pointing out that these statistical features are of vital importance to our MLPR model, which helps our model generalize well to test problem instances that are not seen during training. We will describe these statistical features in Section \ref{Subsection: Statistical Measures}. 

Apart from using problem reduction as a preprocessing step (i.e., removing irrelevant decision variables from a given  problem instance \emph{a priori}), there exist other smart uses of problem reduction techniques. \citet{he2014learning} learned a node pruning policy for branch-and-bound algorithms to heuristically cut off branches that are unlikely to generate a better primal solution.  \citet{li2018combinatorial} estimated a likelihood for each decision variable of whether it belongs to an optimal solution, and used the estimated probabilities to guide a tree search algorithm. \citet{ding2019accelerating} trained a graph convolutional network to predict solution values for binary variables, and used the predicted values to generate a  global inequality constraint to prune the search space. These methods are typically designed for a particular type of solution algorithms. In contrast, our MLPR method is more generally applicable and can be used as a preprocessing technique for any existing solution algorithm.   

\subsection{Support Vector Machine}\label{Subsection: Support Vector Machine}

Consider a binary classification task with $m$ training instances $\mathbb{S}=\{(\bm{f}^i,l^i)\,|\, i = 1,\cdots,m\}$, where $\bm{f}^i$ is the feature vector and $l^i\in\{-1 , 1\}$ is the class label of the $i_{th}$ training instance. A classification algorithm aims to find a decision boundary to separate the positive (label $1$) and negative (label $-1$) training instances as well as possible.

We first assume the positive and negative training instances can be separated by a linear classifier ($h$), parameterized by $(\bm{w}$, $b$):
\begin{equation}
h_{\bm{w},b}(\bm{f}) = \sign(\bm{w}^T\bm{f}+b),
\end{equation}
where $\sign(\bm{w}^T\bm{f}+b)$ is the sign of value $\bm{w}^T\bm{f}+b$. The \emph{geometric margin}  of $(\bm{w},b)$ with respect to a training instance $(\bm{f}^i,l^i)$ is defined as  the distance from $\bm{f}^i$ to the decision boundary  ($\bm{w}^T\bm{f}+b = 0$) in the feature space: 
\begin{equation}
\gamma^i = l^i\Big(\frac{\bm{w}^T}{||\bm{w}||}\bm{f}^i+\frac{b}{||\bm{w}||}\Big). \end{equation}
The geometric margin of $(\bm{w},b)$ with respect to a training set $\mathbb{S}$ is the smallest geometric margin to the individual training instances: 
\begin{equation}
\gamma = \min_{i=1,\cdots,m}\gamma^i.
\end{equation}

The aim of SVM \citep{boser1992training,cortes1995support} is to find a decision boundary, determined by ($\bm{w}, b$), that maximizes the geometric margin $\gamma$:
\begin{align}
\max_{\gamma,\bm{w},b} & \quad \gamma, \\ 
s.t. & \quad l^i\Big(\frac{\bm{w}^T}{||\bm{w}||}\bm{f}^i+\frac{b}{||\bm{w}||}\Big) \ge \gamma, \, i = 1, \cdots m.
\end{align}
SVM is also known as an optimal margin classifier. Scaling $\bm{w}$ and $b$ by any positive number does not change the decision boundary: $\bm{w}\bm{f}+b = 0$. Thus we can restrict the norm of $\bm{w}$ to be any positive number without changing the optimal decision boundary. In order to efficiently solve the optimization problem, the norm of $\bm{w}$ is usually set to $1/\gamma$, i.e., $||\bm{w}||=1/\gamma$. Thus maximizing $\gamma$ is equivalent to maximizing $1/||\bm{w}||$, which is also equivalent to minimizing $\frac{1}{2}\bm{w}^T\bm{w}$. The optimization problem is then transferred to a quadratic programming with linear constraints, which can be solved efficiently:
\begin{align}
\min_{\bm{w},b} & \quad \frac{1}{2}\bm{w}^T\bm{w}, \\ 
s.t. & \quad l^i\big(\bm{w}^T\bm{f}^i+b\big) \ge 1, \, i = 1, \cdots m. 
\end{align}
Let $(\bm{w}^*,b^*)$ determine the optimal decision boundary. The feature vector $\bm{f}^i$ is called a  \emph{support vector} if $l^i\big({\bm{w}^*}^T\bm{f}^i+b^*\big) = 1$. The support vectors are the training instances with the smallest geometric margin (those closest to the optimal decision boundary). Thus only support vectors can influence the optimal decision boundary. Adding or deleting a training instance which is not a support vector does not change the optimal decision boundary. 

%

%
%

When the training set $\mathbb{S}$ cannot be well separated by a linear classifier, we can map the feature vector $\bm{f}$ to a higher-dimensional space using a non-linear function $\phi(\cdot)$, in the hope that the training instances can be separated more easily in the higher-dimensional space. We can also use regularization, that allows a smaller geometric margin at a cost of increasing the objective value. Importantly this also caters for the case where the given set of non-linear functions  is unable to provide a correct classification for all training instances. The \emph{primal} optimization problem becomes
\begin{align}
\min_{\bm{w},b, \bm{\xi}} & \quad \frac{1}{2}\bm{w}^T\bm{w} + r^+ \sum_{l^i=1}g(\xi^i) + r^- \sum_{l^i=-1}g(\xi^i),\label{Eq. SVM_primal}\\ 
s.t. & \quad l^i\big(\bm{w}^T\phi(\bm{f}^i)+b\big) \ge 1 - \xi^i, &\quad& i = 1, \cdots m,\label{Eq. const_primal}\\
& \quad \xi^i \ge 0, && i = 1, \cdots m,
\end{align}
where  $\phi(\bm{f}^i)$ maps the feature vector $\bm{f}^i$ into a higher-dimensional space; $r^+>0$ and $r^->0$ are the regularization parameters for positive and negative training instances; $\xi^i$, $i=1,\cdots,m$ are slack variables and $g(\cdot)$ is a loss function. We will denote SVM with first order loss function $g(\xi^i) := \xi^i$ as L1-SVM, and SVM with second order loss function $g(\xi^i) := (\xi^i)^2$  as L2-SVM.  

If the function $\phi(\bm{f}^i)$ maps the feature vector  $\bm{f}^i$ to a very high dimensional space, solving the primal optimization problem is computationally slow. In this case, the \emph{dual} problem may be easier to solve. Considering L1-SVM, the \emph{dual} optimization problem is 
\begin{align}
\min_{\bm{\alpha}} & \quad \frac{1}{2}\bm{\alpha}^TQ\bm{\alpha} - \bm{e}^T\bm{\alpha},\\
s.t. & \quad \bm{l}^T\bm{\alpha} = 0, \\
\label{Cons: 1} & \quad 0 \le \alpha_i \le r^+, &&  \forall \, i\in\{1,\cdots,m\} \; \text{and} \; l^i = 1, \\
\label{Cons: 2} & \quad 0 \le \alpha_i \le r^-, && \forall \,  i\in\{1,\cdots,m\} \; \text{and} \; l^i = -1,
\end{align}
where $\{\alpha_1, \cdots, \alpha_m$\} are dual variables of constraints (\ref{Eq. const_primal}), $\bm{e} = [1,\cdots, 1]^T$ is the vector of all ones,  $Q$ is an $m\times m$ positive semidefinite matrix, and $Q_{i,j} = l^il^jK(\bm{f}^i,\bm{f}^j)$, and $K(\bm{f}^i,\bm{f}^j) = \phi(\bm{f}^i)\phi(\bm{f}^j)$ is the kernel function. The kernel function avoids the need to explicitly compute $\phi(\cdot)$, thus is computationally efficient.  For example the radial basis function (RBF), defined as $K_{rbf}(\bm{f}^i, \bm{f}^j) = \exp(-\gamma_k||\bm{f}^i-\bm{f}^j||^2)$, where $\gamma_k$ is a kernel parameter, implicitly maps the  feature space to an infinity dimensional space. But computing the RBF kernel only costs $\mathcal{O}(m)$. Let $\alpha_i^*$,  $i=1,\cdots,m$ denote the optimal dual values. Due to the KKT dual complementarity condition, if $\alpha_i^*>0$ the corresponding training instance $\bm{f}^i$ is a support vector. As $\bm{w}^* = \sum_{i=1}^{m}\alpha_i^*l^i\bm{f}^i$, the optimal decision boundary and thus the prediction for a given new instance are only determined by the support vectors (those with $\alpha_i^*>0$). It is noteworthy that the number of support vectors is usually much smaller than the number of training instances in the training set. 

\section{Problem Reduction for Travelling Salesman Problem Using Machine Learning}\label{Section: MLPR}
In this section, we adapt our MLPR method originally proposed in \citep{sun2019using} to prune the search space for TSP. We model TSP as a complete graph $G(V,E,C)$, where $V$ denotes a set of cities, $E$ denotes edges between cities, and $C$ denotes edge costs (e.g., distance between cities). The objective of TSP is to search for a  route with minimum edge costs that visits each vertex and return to the original vertex. Our MLPR method uses machine learning to predict for each edge whether it belongs to an optimal route, and removes from the complete graph the edges that are not expected to be part of an optimal route. 

We use optimally-solved TSP instances as our training set, and treat each edge in a solved graph as a training instance. We assign a class label $1$ to the edges that belong to the optimal route and $-1$ to those who do not. We will extract two statistical measures and four graph features to characterize each edge in Section \ref{Subsection: Features}. After constructing a training set, our goal is then to learn a decision boundary in the feature space to differentiate between positive (with class label $1$) and negative (with class label $-1$) training instances. This becomes a typical binary classification problem and any classification algorithm can be used for this task. We will use SVM to learn a decision boundary for this task in Section \ref{Subsection: SVM}. For a given large TSP instance where we do not know the optimal route, the trained model can then be used to predict a class label for each edge in the graph. By removing the edges that are predicted to be $-1$, we have a reduced sparse graph, which is hopefully much easier for an existing solution algorithm to solve. The main steps of our MLPR method for TSP are summarized in Algorithm \ref{Algorithm: MLPR}. 
\begin{algorithm}[!t]
	\caption{\textsc{MLPR for TSP}}
	\label{Algorithm: MLPR}
	\begin{algorithmic}[1]
		\State Solve selected easy TSP instances to optimality;
		\State Assign a class label $1$ to the edges in the optimal route and $-1$ to others;
		\State Extract features to characterize each edge (training instance); 
		\State Train a machine learning model to separate positive and negative edges; 
		\State Predict a class label for each edge on an unseen test graph (where its optimal solution is unknown) using the trained model, and remove negative edges. 				
	\end{algorithmic}
\end{algorithm}

\subsection{Extracting Features to Characterize Each Edge}\label{Subsection: Features}
We extract four features directly computed from graph data and two statistical measures computed from stochastic samples of feasible solutions to characterize each edge (training instance).  

\subsubsection{Graph Features}
As the objective of TSP is to search for a route with minimum costs, the edge cost is an important feature related to the objective value. Considering a TSP instance $G(V,E,C)$ with $n$ cities, we design four graph features to describe each edge  $e_{i,j}$, $i, j = 1,\cdots, n$, based on the edge costs $C$: 
\begin{equation}
f_1(e_{i,j}) = \frac{c_{i,j} - \min\limits_{k = 1, \cdots, n} c_{i,k}}{\max\limits_{k = 1, \cdots, n} c_{i,k} - \min\limits_{k = 1, \cdots, n} c_{i,k}},
\end{equation} 
\begin{equation}
f_2(e_{i,j}) = \frac{c_{i,j} - \min\limits_{k = 1, \cdots, n} c_{k,j}}{\max\limits_{k = 1, \cdots, n} c_{k,j} - \min\limits_{k = 1, \cdots, n} c_{k,j}},
\end{equation}
\begin{equation}
f_3(e_{i,j}) = \frac{c_{i,j} - \sum_{k=1}^{n} c_{i,k}/n}{\max\limits_{k = 1, \cdots, n} c_{i,k} - \min\limits_{k = 1, \cdots, n} c_{i,k}},
\end{equation} 
\begin{equation}
f_4(e_{i,j}) = \frac{c_{i,j} - \sum_{k=1}^{n} c_{k,j}/n}{\max\limits_{k = 1, \cdots, n} c_{k,j} - \min\limits_{k = 1, \cdots, n} c_{k,j}}.
\end{equation}
The first feature computes the difference between the edge cost of $e_{i,j}$ ($c_{i,j}$) and the minimum edge cost that originates from vertex $i$, while the second feature computes the difference between $c_{i,j}$ and the minimum edge cost that ends in vertex $j$. The third and fourth features are computed based on the mean edge costs connected to vertex $i$ or $j$. We normalize the four features by the difference between the maximum and minimum edge costs  that connect to vertex $i$ or $j$. These graph features only capture local characteristics of an edge. In the next subsection, we will describe two statistical measures to capture certain global features for each edge. 


\subsubsection{Statistical Measures}\label{Subsection: Statistical Measures}
The statistical measures aim to quantify the likelihood of each edge belonging to an optimal route based on randomly generated samples of feasible routes. As TSP has the backbone structure \citep{kilby2005backbone}, it is possible to identify the edges shared between an optimal route and high-quality routes. 

Randomly generating a  feasible route for TSP  is very simple. Supposing the vertices (cities) are labelled from $1$ to $n$, any random permutation ($P$) of integers from $1$ to $n$ is a feasible route for visiting each city. We generate $m$ random feasible routes $\{P^1,P^2,\cdots,P^m\}$, and compute the corresponding  objective values $\{y^1,y^2,\cdots,y^m\}$. The time complexity of sampling is $\Theta(mn)$, simply because a random permutation of $n$ elements costs $\Theta(n)$. 

To define our statistical measures, we introduce a binary string $\bm{x}^k$ to represent the $k_{th}$ sample route $P^k$, where $x_{i,j}^k = 1$ means the edge $e_{i,j}$ is in the $k_{th}$ sample; otherwise it is not. The first statistical measure is computed from the ranking of sample routes. We sort the sample routes based on their objective values in ascending order, and use  $r^k$ to denote the ranking of the $k_{th}$ sample. The ranking-based measure for edge $e_{i,j}$ is defined as 
\begin{equation}\label{Eq: ranking-based measure}
f_r(e_{i,j}) = \sum_{k=1}^{m}\frac{x_{i,j}^k}{r^k},
\end{equation}
where $i, j = 1, \cdots, n$. The edges that frequently appear in high-quality sample routes have a large ranking-based score, and are more likely to be part of an optimal route. We then normalize each ranking-based score by dividing the maximum ranking-based score in a graph
\begin{equation}
f_5(e_{i,j}) =\frac{f_r(e_{i,j})}{\max\limits_{p,q = 1,\cdots,n} f_r(e_{p,q})}. 
\end{equation}
This normalization avoids a large-valued feature dominating a classification task. 

The second statistical measure we have developed is a correlation-based measure, that computes the Pearson correlation coefficient between each variable $x_{i,j}$ and objective values across the sample routes:
\begin{equation}\label{Eq: correlation-based measure}
f_c(e_{i,j}) = \frac{\sum_{k=1}^{m}(x_{i,j}^k - \bar{x}_{i,j})(y^k - \bar{y}) }{\sqrt{\sum_{k=1}^{m}(x_{i,j}^k-\bar{x}_{i,j})^2}\sqrt{\sum_{k=1}^{m}(y^k-\bar{y})^2}},
\end{equation}
where $\bar{x}_{i,j} = \sum_{k= 1}^{m}x_{i,j}^k/m$, and $\bar{y} = \sum_{k=1}^{m}y^k/m$. As TSP is a minimization problem, edges that are highly negatively correlated with the objective values are likely to be in an optimal route. Similarly, we normalize the correlation-based score by the minimum correlation value in a graph:
\begin{equation}
f_6(e_{i,j}) = \frac{f_c(e_{i,j})}{\min\limits_{p,q = 1, \cdots, n} f_c(e_{p,q})}. 
\end{equation}

Directly computing these two statistical measures from the binary string representation $\bf{x}$ costs $\Theta(mn^2)$ in both time and space complexity. In \citep{sun2019using}, we have introduced an efficient method based on set representation, i.e., permutation $P$ here. Because $x_{i,j}^k$ are binary variables, we can simplify the calculation of Pearson correlation coefficient using the following two equalities:
\begin{equation}
\sum_{k=1}^{m}(x_{i,j}^k-\bar{x}_{i,j})^2 = \bar{x}_{i,j}(1-\bar{x}_{i,j})m,
\end{equation}
\begin{equation}
\sum_{k=1}^{m}(x_{i,j}^k - \bar{x}_{i,j})(y^k - \bar{y})  = (1- \bar{x}_{i,j}) s_{i,j}^1 - \bar{x}_{i,j} s_{i,j}^0,
\end{equation}
where 
\begin{equation}
s_{i,j}^1 = \mathop{\sum\limits_{1\le k \le m}}\limits_{x_{i,j}^k = 1}(y^k - \bar{y}); \; \text{and} \; s_{i,j}^0 = \mathop{\sum\limits_{1\le k \le m}}\limits_{x_{i,j}^k = 0}(y^k - \bar{y}). 
\end{equation}
The proof of these two equalities can be found in \citep{sun2019using}. Having this simplification, we can compute the two statistical measures in $\Theta(mn+n^2)$ for both time and space complexity using Algorithm \ref{Algorithm: Statistical Measures}. The main idea is to iterate through the edges in each sample route $P$ to accumulate our ranking-based measure $f_r(e_{i,j})$, $\bar{x}_{i,j}$ and $s_{i,j}^1$, i.e., line \ref{line: for1} to \ref{line: for2} in Algorithm \ref{Algorithm: Statistical Measures}.  Our correlation-based measure $f_c(e_{i,j})$ can then be easily computed based on $\bar{x}_{i,j}$ and $s_{i,j}^1$. 

In practice, the sample size $m$ should be larger than $n$; otherwise there will be some edges that are never sampled. Considering a symmetric TSP instance with $n$ cities, the number of edges in the undirected complete graph is $n(n-1)/2$. The total number of edges in $m$ sample routes is $mn$. Thus each edge is expected to be sampled $2m/(n-1)$ times. In our experiments, we set $m = 100n$. 

It is noteworthy that the two statistical measures described here can be directly used as a problem reduction technique, e.g., we can remove the edges that are positively correlated with objective values from a graph. However, as we have shown in \citep{sun2019using}, the machine learning approach which takes these two statistical measures as features outperforms a single statistical measure for problem reduction. Thus we will simply use both statistical measures as inputs to our machine learning model in this paper. 

\begin{algorithm}[!t]
	\caption{\textsc{Statistical Measures}($\mathbb{P}$, $Y$, $m$, $n$)}
	\label{Algorithm: Statistical Measures}
	\begin{algorithmic}[1]
		\State Sort the samples in $\mathbb{P}$ based on objective value $Y$; use $r^k$ to denote the ranking of $k_{th}$ sample $P^k$;
		\State Compute mean objective value: $\bar{y} \leftarrow \sum_{k=1}^{m}y^k/m$;
		\State Compute objective difference: $y_d \leftarrow \sum_{k=1}^{m}(y^k - \bar{y})$; 
		\State Compute objective variance: $\sigma_y \leftarrow \sum_{k=1}^{m}(y^k - \bar{y})^2$; 
		\State Initialize $f_r$, $\bar{x}_{i,j}$ and $s_{i,j}^1$ to $0$,  for each $e_{i,j} \in E$; 					
		\For{$k$ from $1$ to $m$}\label{line: for1}
			\For{$idx$ from $1$ to $n$}
				\If{$idx < n$}
				\State $i \leftarrow P^k[idx]$, $j \leftarrow P^k[idx+1]$;
				\Else
				\State $i \leftarrow P^k[idx]$, $j \leftarrow P^k[1]$;
				\EndIf
				\State $f_r(e_{i,j}) \leftarrow f_r(e_{i,j}) + 1/r^k$;
				\State $\bar{x}_{i,j} \leftarrow \bar{x}_{i,j} + 1 / m$;
				\State $s_{i,j}^1 \leftarrow s_{i,j}^1 + (y^k - \bar{y})$;
			\EndFor
		\EndFor\label{line: for2}			
		\For{$i$ from $1$ to $n$}
		\For{$j$ from $1$ to $n$ and $j\ne i$}
		\State $\sigma_{c_{i,j}} \leftarrow (1-\bar{x}_{i,j})s_{i,j}^1 - \bar{x}_{i,j}(y_d - s_{i,j}^1)$;
		\State $\sigma_{x_{i,j}}  \leftarrow \bar{x}_{i,j}(1-\bar{x}_{i,j})m$;
		\State $f_c(e_{i,j}) \leftarrow  \sigma_{c_{i,j}}/ \sqrt{\sigma_{x_{i,j}}\sigma_y}$; 
		\EndFor
		\EndFor\\
		\Return $f_r$ and $f_c$. 
	\end{algorithmic}
\end{algorithm}

\subsection{Support Vector Machine Classification}\label{Subsection: SVM}

In our training set, the number of positive training instances is much smaller than that of negative instances. Considering a symmetric TSP instance with $n$ cities, the number of edges in an optimal route is $n$, and the total number of edges is $n(n-1)/2$. Thus the ratio between positive and negative edges is $2:(n-3)$. The standard SVM formulation tends to classify negative training instances better than the positive instances, because there are more negative training instances. However, misclassifying a positive instance is much more harmful than misclassifying a negative instance. If a positive instance is misclassified, the reduced optimization problem no longer captures the original optimal solution. On the other hand, misclassifying a negative instance only results in a slight increase of the reduced problem size. In this sense, we will use the cost-sensitive SVM (see Section \ref{Subsection: Support Vector Machine}) and penalize misclassifying positive instances more by using a larger regularization parameter $r^+$ in Eq. (\ref{Eq. SVM_primal}), in contrast to that of negative instances $r^-$. In our experiments, we will set $r^- = 1$ and $r^+ = \epsilon_m n_{-1}/n_1$, where $n_{-1}$ and $n_1$ are the number of negative and positive instances in our training set, and $\epsilon_m$ controls the penalty for misclassifying positive instances. The term $n_{-1}/n_1$ balances the number of positive and negative instances in our training set. 

We will consider two types of SVM in our experiments, linear SVM (solving primal optimization problem) and non-linear SVM with the RBF kernel (solving dual optimization problem); see Section~\ref{Subsection: Support Vector Machine} for more details. The classification accuracy of kernel SVM is usually higher than that of linear SVM. We will use the SMO-type (Sequential Minimal Optimization) decomposition method \citep{fan2005working} implemented in the LIBSVM library  \citep{chang2011libsvm}  to solve the dual optimization problem of L1-SVM. However when the number of training instances is too large (e.g, millions of instances), solving the dual problem is computationally very slow. In this case, we will solve the primal optimization problem of linear L2-SVM using the trust region Newton method \citep{lin2008trust} implemented in the LIBLINEAR library \citep{fan2008liblinear}. 

\section{Experiments}
\label{Section: Experimental Results}

In this section, we use simulation experiments to investigate the robustness of our MLPR model. We will consider three scenarios where training and test instances are different, and explore the corresponding generalization errors. Specifically in Section~\ref{Subsection: Generalization Regarding Problem Characteristics}, we train our MLPR model using one category of TSP instances and test it on another with different problem characteristics. In Section~\ref{Subsection: Generalization Regarding Problem Size}, we train MLPR using small randomly generated TSP instances, and test it on larger randomly generated TSP instances. In Section~\ref{Subsection: Generalization Regarding Problem Variants}, we train our MLPR model on symmetric TSP instances, and test it on three TSP variants. In Section \ref{Subsection: Boosting the Performance of CPLEX}, we investigate whether our MLPR method can be used as a preprocessing technique to boost the performance of a generic solver -- CPLEX. In the last subsection, we compare our MLPR method against other generic problem reduction methods. Our source codes are implemented in C++, and compiled with GCC/7.3.0-2.30. All our experiments are conducted on a high performance computing server with Intel(R) Xeon(R) Gold 6154 CPUs @ 3.00 GHz and 21 GB RAM.\footnote{Our C++ source codes are publicly available online at \url{https://github.com/yuansuny/tsp}.} 



\subsection{Varying Problem Characteristics}\label{Subsection: Generalization Regarding Problem Characteristics}

\subsubsection{Setting} 

We use the TSP instances from the MATILDA library as our dataset\footnote{https://matilda.unimelb.edu.au/matilda/}. This library contains 7 categories, each with 190 symmetric TSP instances. The instances in each category are evolved by a genetic algorithm to have certain problem characteristics, such that they are hard (or easy) for a particular heuristic to solve \citep{smith2011discovering}. They have considered two heuristic methods -- Chained Lin-Kernighan (CLK) \citep{applegate2003chained} and Lin-Kernighan with Cluster Compensation (LKCC) \citep{johnson1997traveling}, resulting in 7 categories of instances: CLKeasy ($\mathcal{I}_1$), CLKhard ($\mathcal{I}_2$), easyCLK-hardLKCC ($\mathcal{I}_3$), hardCLK-easyLKCC ($\mathcal{I}_4$), LKCCeasy ($\mathcal{I}_5$), LKCChard ($\mathcal{I}_6$) and random ($\mathcal{I}_7$).

Because the dimensionality of these instances is small (i.e., 100), we can quickly solve these instances to optimality using the Concorde solver \citep{applegate2006concorde}. This enables us to systematically evaluate the generalization capability of our MLPR model to instances in different categories. We train a MLPR model using the first 50 instances in one category, and test the trained model on the remaining unseen 140 instances in that category as well as the instances from other 6 categories. For each test instance, we apply the trained model to reduce the problem size (i.e., pruning some edges in the complete graph), and solve the reduced problem to optimality using Concorde.\footnote{We do not remove the edges that appear in the best sample solution to guarantee that the reduced problem space contains at least one feasible solution.} We compute an optimality gap by comparing the optimal solutions generated in the reduced and original problems. As our MLPR model uses statistical features computed from random samples, the reduced problem generated for a test instance might be slightly different if we use a different random seed. Therefore, we repeat the random sampling and problem reduction process 25 times to alleviate randomness. 



As the size of each training set is not too large, we use L1-SVM with RBF kernel to train our MLPR model. The computational time required for training one model is less than 30 minutes, and the predicting time for one test instance is around 5 seconds. The kernel parameter $\gamma_k$ is set to the default value used in the LIBSVM library: $\gamma_k = 1/n_f$, where $n_f$ is the number of features. We have tested multiple values for the penalty parameter $\epsilon_m$ in \citep{sun2019using}, and found that $\epsilon_m = 10$ works reasonably well across a wide range of problem instances. Thus we simply set $\epsilon_m = 10$ in this subsection.



\subsubsection{Results}
The average optimality gap generated by our MLPR model and remaining problem size after reduction are presented in Tables \ref{Table: the optimality gap} and \ref{Table: the percentage of edges}, respectively. These two tables should be read in conjunction as, in an ideal world, we would want to see both small reduced problems and small optimality gaps. What the results show in practice is that, unsurprisingly, the MLPR models that retain a larger fraction of the original problem also tend to produce smaller gaps. Hence, the best optimality gaps in Table~\ref{Table: the optimality gap} do not simply occur on the diagonal as one might first expect. As a compromise between these two aims of minimising gaps and size, we observe that using randomly generated TSP instances ($\mathcal{I}_7$) as a training set, our MLPR model performs reasonably well across all 7 categories of TSP instances. The MLPR-$\mathcal{I}_7$ model prunes about 85\% of edges, and overall achieves 0.44\% optimality gap. Further, the TSP instances that are hard for CLK or LKCC to solve (e.g., $\mathcal{I}_2$ and $\mathcal{I}_6$) are also hard for our MLPR model. The optimality gaps generated by our MLPR models for hard TSP instances are larger than those for easy TSP instances.  Third, MLPR trained on easy TSP instances (e.g., MLPR-$\mathcal{I}_1$ or MLPR-$\mathcal{I}_5$) prunes too many edges for hard instances (e.g., $\mathcal{I}_2$ or $\mathcal{I}_6$), resulting in a large optimality gap. On the other hand, MLPR trained on hard TSP instances prunes too few edges for the easy instances tested, resulting in larger problem sizes still remaining. For example, both MLPR-$\mathcal{I}_2$ and MLPR-$\mathcal{I}_3$ achieve $0$ optimality gap for easy instances in $\mathcal{I}_1$, but MLPR-$\mathcal{I}_2$ prunes 8.5\% few edges than MLPR-$\mathcal{I}_3$. Lastly, we observe that when training and test instances are from the same category, the generated optimality gap is less than $0.2\%$, and the amount of pruned edges is reasonable -- in fact the results are always non-dominated in the Pareto sense, with all other models producing worse gaps or larger remaining problems.

\begin{table}[!t]
	\centering
	\caption{The average optimality gap (\%) generated by our MLPR model when trained on one category of TSP instances and tested on another. The 7 categories of instances are labelled as $\mathcal{I}_1, \cdots, \mathcal{I}_7$. The MLPR model trained on category $\mathcal{I}_j$ is denoted as MLPR-$\mathcal{I}_j$. The best optimality gap generated for each test category is in bold.}
	\label{Table: the optimality gap}
	\begin{tabular}{l l l l l l l l}
		\toprule
		Models & $\mathcal{I}_1$ & $\mathcal{I}_2$ & $\mathcal{I}_3$ & $\mathcal{I}_4$ & $\mathcal{I}_5$ & $\mathcal{I}_6$ & $\mathcal{I}_7$ \\\midrule
		MLPR-$\mathcal{I}_1$ & 0.11 & 11.19 & 4.43 & 5.57 & 0.90 & 6.95 & 4.06 \\
		MLPR-$\mathcal{I}_2$ & \bf{0.00} & \bf{0.11}  & 0.03 & \bf{0.00} & \bf{0.00} & 0.40 & \bf{0.00} \\
		MLPR-$\mathcal{I}_3$ & \bf{0.00} & 2.44  & 0.18 & 0.25 & \bf{0.00} & 1.18 & 0.09 \\
		MLPR-$\mathcal{I}_4$ & \bf{0.00} & 1.94  & 0.18 & 0.14 & 0.02 & 1.04 & 0.07 \\
		MLPR-$\mathcal{I}_5$ & 0.04 & 8.38  & 1.83 & 2.60 & 0.17 & 3.87 & 1.41 \\
		MLPR-$\mathcal{I}_6$ & \bf{0.00} & 0.12  & \bf{0.00} & \bf{0.00} & \bf{0.00} & \bf{0.18} & \bf{0.00} \\
		MLPR-$\mathcal{I}_7$ & \bf{0.00} & 1.75  & 0.19 & 0.10 & \bf{0.00} & 1.01 & 0.05 \\\bottomrule
	\end{tabular}
\end{table}

\begin{table}[!t]
	\centering
	\caption{The percentage of remaining problem size with respect to its original problem size (\%) when training our MLPR model on one category of TSP instances and testing it on another. The 7 categories of instances are labelled as $\mathcal{I}_1, \cdots, \mathcal{I}_7$. The MLPR model trained on category $\mathcal{I}_j$ is denoted as MLPR-$\mathcal{I}_j$.}
	\label{Table: the percentage of edges}
	\begin{tabular}{l l l l l l l l}
		\toprule
		Models & $\mathcal{I}_1$ & $\mathcal{I}_2$ & $\mathcal{I}_3$ & $\mathcal{I}_4$ & $\mathcal{I}_5$ & $\mathcal{I}_6$ & $\mathcal{I}_7$ \\\midrule
		MLPR-$\mathcal{I}_1$ & 8.30  & 9.15  & 8.66  & 8.57  & 8.14  & 9.33  & 8.33  \\
		MLPR-$\mathcal{I}_2$ & 22.24 & 21.49 & 23.67 & 21.67 & 21.80 & 23.35 & 22.20 \\
		MLPR-$\mathcal{I}_3$ & 13.79 & 13.22 & 14.22 & 13.47 & 13.42 & 14.56 & 13.66 \\
		MLPR-$\mathcal{I}_4$ & 14.47 & 14.12 & 15.70 & 14.07 & 14.05 & 16.00 & 14.46 \\
		MLPR-$\mathcal{I}_5$ & 9.68  & 10.10 & 9.97  & 9.76  & 9.44  & 10.61 & 9.63  \\
		MLPR-$\mathcal{I}_6$ & 23.50 & 21.76 & 23.36 & 23.11 & 23.23 & 22.80 & 23.00 \\
		MLPR-$\mathcal{I}_7$ & 14.94 & 14.35 & 15.71 & 14.53 & 14.56 & 15.99 & 14.82 \\\bottomrule
	\end{tabular}
\end{table}

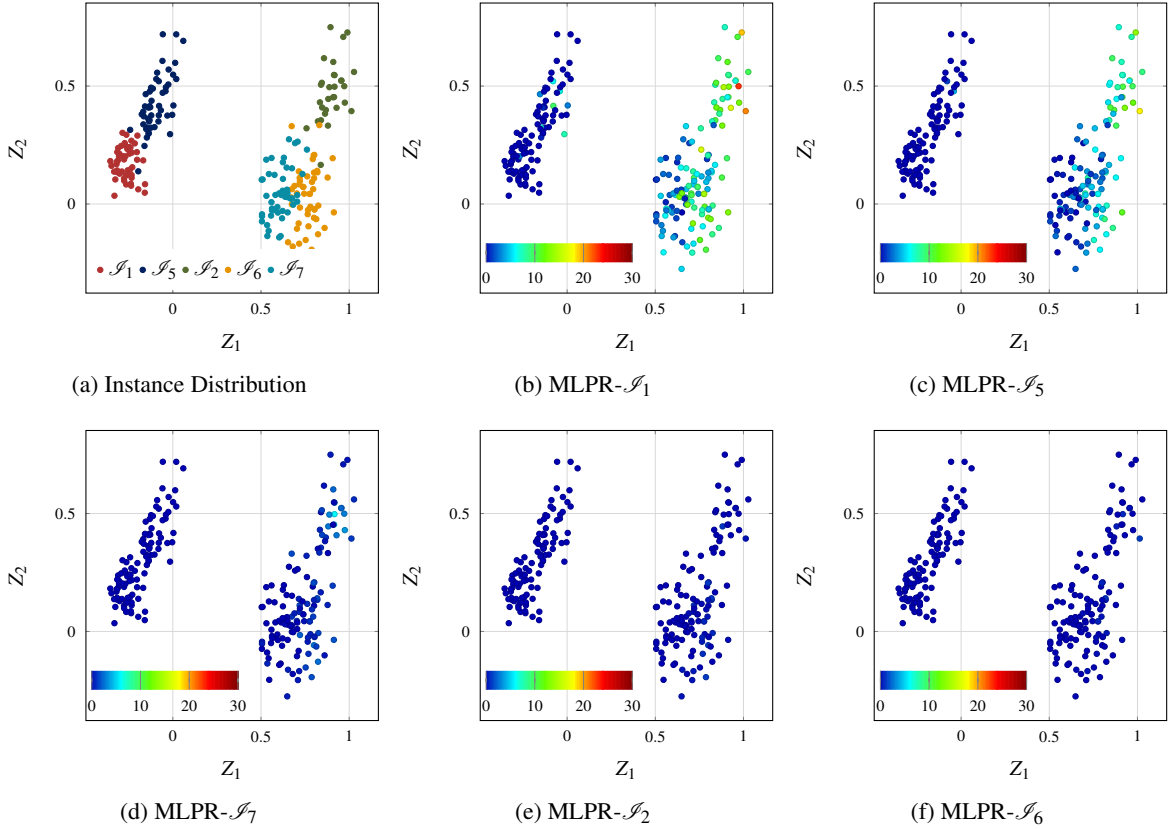
\begin{figure}[!t]
	\subfloat[Instance Distribution]{
		\label{Subfig: Instance Distribution}
		\begin{tikzpicture}
		\begin{axis}[box plot width=0.20em, xlabel = \scriptsize $Z_1$, ylabel = \scriptsize $Z_2$, height=0.33\textwidth,width=0.33\textwidth, grid style={line width=.1pt, draw=gray!10},major grid style={line width=.2pt,draw=gray!30}, xmajorgrids=true, ymajorgrids=true,  major tick length=0.05cm, minor tick length=0.0cm, legend style={at={(0.01,0.08)},anchor=west,font=\scriptsize,draw=none,legend columns=-1}]
		\addplot[
		scatter,only marks,scatter src=explicit symbolic, mark size =1pt,
		scatter/classes={
			1={mark=*,maroon},
			2={mark=*,marine},
			3={mark=*,olivegreen},
			4={mark=*,orange},
			5={mark=*,blue}
		}
		]
		table[x=x,y=y,meta=label]{\CLKeasy};
		\legend{$\mathcal{I}_1$,$\mathcal{I}_5$,$\mathcal{I}_2$,$\mathcal{I}_6$,$\mathcal{I}_7$}
		\end{axis}
		\end{tikzpicture}
	}
	\subfloat[MLPR-$\mathcal{I}_1$]{
		\label{Subfig: MLPR-I1}
		\begin{tikzpicture}
		\begin{axis}[box plot width=0.20em, xlabel = \scriptsize $Z_1$, ylabel = \scriptsize $Z_2$, height=0.33\textwidth,width=0.33\textwidth, grid style={line width=.1pt, draw=gray!10},major grid style={line width=.2pt,draw=gray!30}, xmajorgrids=true, ymajorgrids=true,  major tick length=0.05cm, minor tick length=0.0cm, legend style={at={(0.05,0.80)},anchor=west,font=\scriptsize,draw=none}, colormap/bluered, point meta max = 30, point meta min = 0,
		colorbar horizontal, colorbar style={at={(0.02,0.0)}, anchor=below south west, font=\scriptsize, axis line style={draw=none}, width=0.5*\pgfkeysvalueof{/pgfplots/parent axis width},},colorbar/width=2.5mm]
		\addplot[scatter,only marks, mark size =1pt, scatter src=explicit]table[x=x,y=y,meta=error]{\CLKeasy};
		\end{axis}
		\end{tikzpicture}
	}
	\subfloat[MLPR-$\mathcal{I}_5$]{
		\begin{tikzpicture}
		\begin{axis}[box plot width=0.20em, xlabel = \scriptsize $Z_1$, ylabel = \scriptsize $Z_2$, height=0.33\textwidth,width=0.33\textwidth, grid style={line width=.1pt, draw=gray!10},major grid style={line width=.2pt,draw=gray!30}, xmajorgrids=true, ymajorgrids=true,  major tick length=0.05cm, minor tick length=0.0cm, legend style={at={(0.05,0.80)},anchor=west,font=\scriptsize,draw=none}, colormap/bluered, point meta max = 30, point meta min = 0, colorbar horizontal, colorbar style={at={(0.02,0.0)}, anchor=below south west,font=\scriptsize, axis line style={draw=none}, width=0.5*\pgfkeysvalueof{/pgfplots/parent axis width},},colorbar/width=2.5mm]
		\addplot[ scatter,only marks, mark size =1pt, scatter src=explicit]table[x=x,y=y,meta=error]{\LKCCeasy};
		\end{axis}
		\end{tikzpicture}
	}
	
	\subfloat[MLPR-$\mathcal{I}_7$]{
		\begin{tikzpicture}
		\begin{axis}[box plot width=0.20em, xlabel = \scriptsize $Z_1$, ylabel = \scriptsize $Z_2$, height=0.33\textwidth,width=0.33\textwidth, grid style={line width=.1pt, draw=gray!10},major grid style={line width=.2pt,draw=gray!30}, xmajorgrids=true, ymajorgrids=true,  major tick length=0.05cm, minor tick length=0.0cm, legend style={at={(0.05,0.80)},anchor=west,font=\scriptsize,draw=none}, colormap/bluered, point meta max = 30, point meta min = 0, colorbar horizontal, colorbar style={at={(0.02,0.0)}, anchor=below south west,font=\scriptsize, axis line style={draw=none}, width=0.5*\pgfkeysvalueof{/pgfplots/parent axis width},},colorbar/width=2.5mm]
		\addplot[ scatter,only marks, mark size =1pt, scatter src=explicit]table[x=x,y=y,meta=error]{\random};
		\end{axis}
		\end{tikzpicture}
	}
	\subfloat[MLPR-$\mathcal{I}_2$]{
		\begin{tikzpicture}
		\begin{axis}[box plot width=0.20em, xlabel = \scriptsize $Z_1$, ylabel = \scriptsize $Z_2$, height=0.33\textwidth,width=0.33\textwidth, grid style={line width=.1pt, draw=gray!10},major grid style={line width=.2pt,draw=gray!30}, xmajorgrids=true, ymajorgrids=true,  major tick length=0.05cm, minor tick length=0.0cm, legend style={at={(0.05,0.80)},anchor=west,font=\scriptsize,draw=none}, colormap/bluered, point meta max = 30, point meta min = 0, colorbar horizontal, colorbar style={at={(0.02,0.0)}, anchor=below south west,font=\scriptsize, axis line style={draw=none}, width=0.5*\pgfkeysvalueof{/pgfplots/parent axis width},},colorbar/width=2.5mm]
		\addplot[ scatter,only marks, mark size =1pt, scatter src=explicit]table[x=x,y=y,meta=error]{\CLKhard};
		\end{axis}
		\end{tikzpicture}
	}
	\subfloat[MLPR-$\mathcal{I}_6$]{
		\label{Subfig: MLPR-I6}
		\begin{tikzpicture}
		\begin{axis}[box plot width=0.20em, xlabel = \scriptsize $Z_1$, ylabel = \scriptsize $Z_2$, height=0.33\textwidth,width=0.33\textwidth, grid style={line width=.1pt, draw=gray!10},major grid style={line width=.2pt,draw=gray!30}, xmajorgrids=true, ymajorgrids=true,  major tick length=0.05cm, minor tick length=0.0cm, legend style={at={(0.05,0.80)},anchor=west,font=\scriptsize,draw=none}, colormap/bluered, point meta max = 30, point meta min = 0, colorbar horizontal, colorbar style={at={(0.02,0.0)}, anchor=below south west,font=\scriptsize, axis line style={draw=none}, width=0.5*\pgfkeysvalueof{/pgfplots/parent axis width},},colorbar/width=2.5mm]
		\addplot[ scatter,only marks, mark size =1pt, scatter src=explicit]table[x=x,y=y,meta=error]{\LKCChard};
		\end{axis}
		\end{tikzpicture}
	}
	\caption{The footprint of our MLPR models when tested on different categories of TSP instances. Each dot represents a TSP instance in the 2-dimensional feature space ($Z_1$ and $Z_2$). In figure (a), dot color represents the category where instance is from; while in figure (b) to (f), dot color represents the optimality gap (\%) generated by our MLPR models for the corresponding instance.}
\end{figure}

We use the tool from the MATILDA library to visualize the performance of our MLPR models. Here each TSP instance (instead of an edge) is mapped to a point in a 6-dimensional feature space. The feature values are taken from the MATILDA library and are normalized to the range of 0 to 1. We then apply principal component analysis to reduce the 6-dimensional feature space to 2-dimensional by selecting the first two principal components ($Z_1$ and $Z_2$). The distribution of the TSP instances in the 2-dimensional space, spanned by $Z_1$ and $Z_2$, is shown in Figure \ref{Subfig: Instance Distribution}. For better visualization, we only plot the last 50 TSP instances from each category, and the instances from different categories are in different colors. We can see that instances from different categories are well separated in the feature space. Note that we only visualize five categories of TSP instances, because the feature data for the other two categories is not available in MATILDA. 

We also plot the optimality gap (\%) generated by our MLPR models for these instances when trained on one category of instances at a time in Figure \ref{Subfig: MLPR-I1}-\ref{Subfig: MLPR-I6}. The optimality gap is indicated by the color of the dots (blue is small and red is large). We can observe that the MLPR model trained on easy TSP instances (MLPR-$\mathcal{I}_1$ or MLPR-$\mathcal{I}_5$) does not perform well on hard instances ($\mathcal{I}_2$ or $\mathcal{I}_6$). There appears to be a strong correlation between the size of the gap and the distance from the training data for these instances. The MLPR model trained on hard instances is able to generate a small optimality gap for any instance considered, though this comes at a price of an increased problem size after the reduction.


Finally, we note that although our MLPR method is able to aggressively reduce the problem size of a TSP instance, it does not speed up the specialized Concorde solver as a preprocessing technique to solve the problem. This is because the Concorde solver does not make use of the sparsity of a graph when solving a reduced problem instance, as it transfers a sparse graph to a complete graph by assigning an arbitrary large weight to the edges that do not exist. Here, we further investigate whether our MLPR method can be used as a preprocessing technique to speed up a generic solver CPLEX. To do so, we select six easy problem instances from the TSP library \citep{reinelt1991tsplib}, whose dimension varies between sizes of $40$ and $60$.  We apply the MLPR-$\mathcal{I}_6$ model to prune edges for each test instance, and use CPLEX to optimally solve the original and reduced  instances with default parameter settings. The ratio between the time taken to solve the original and reduced TSP instances is shown in Table \ref{Table: The ratio between the time taken by CPLEX to solve the original and reduced problem instances.}. We can observe that by using our MLPR method as a preprocessing technique, CPLEX achieves 2.68 times of speed-up when solving the 6 test instances on average. Furthermore, the optimal solutions generated from the original and reduced TSP instances are the same, meaning our MLPR method is very accurate at pruning irrelevant edges (i.e., those do not belong to an optimal solution). 

\begin{table}[!t]
	\centering
	\caption{The ratio between the time taken by CPLEX to solve the original and reduced problem instances. This measures how significantly our MLPR-$\mathcal{I}_6$ method speeds up CPLEX as a preprocessing technique.}
	\label{Table: The ratio between the time taken by CPLEX to solve the original and reduced problem instances.}
	\begin{tabular}{lllllll}
		\toprule
		Dataset & Att48 & Berlin52 & Eil51 & Gr48 & Hk48 & Swiss42\\
		Speed-up & 1.93 & 3.92 &  1.96 & 2.74 & 1.78 & 3.01\\\bottomrule
	\end{tabular}
\end{table}

\subsection{Varying Problem Size}\label{Subsection: Generalization Regarding Problem Size}

\subsubsection{Setting}

In this subsection, we explore the generalization error of our MLPR model in terms of problem size. We train our MLPR model using 190 randomly generated TSP instances ($\mathcal{I}_7$) from the MATILDA library with dimension 100, and test it on larger randomly generated TSP instances. Each TSP instance in $\mathcal{I}_7$ of MATILDA is created by randomly generating 100 pairs of integer coordinates between 0 and 400. We use the same method to create test instances with different number of cities, i.e., 200, 500, 800, 1100, 1400, 1700, and 2000. This results in 7 categories, each with 190 randomly generated TSP instances. Because each edge (instead of a complete graph) is a training instance, our training set size is close to 1 million. Thus we use a linear SVM with $\epsilon_m = 10$ to train our MLPR model, by solving the primal problem to gain computational efficiency. We apply the trained model to reduce the problem size for each test TSP instance, and solve the reduced problem as well as the original problem to optimality using Concorde. 


\subsubsection{Results}
\begin{figure}[!t]
\centering
\begin{tikzpicture}
	\begin{axis} [box plot width=0.20em, xtick=data, ymin = 0, ymax= 15, ylabel = \scriptsize$\bar{p}$ (\%),  height=0.40\textwidth,width=0.50\textwidth, grid style={line width=.1pt, draw=gray!10},major grid style={line width=.2pt,draw=gray!30}, xmajorgrids=true, ymajorgrids=true,  major tick length=0.05cm, minor tick length=0.0cm, legend style={at={(0.65,0.40)},anchor=west,font=\scriptsize,draw=none}]
	\addplot[color=marine,mark=*, mark size = 1.5, line width=0.40mm] table{\TSPSizeP}; 				
	\end{axis}
\end{tikzpicture}
\caption{The percentage of remaining problem size after reduction when training our MLPR model on small TSP instances and testing it on larger instances. The horizontal axis represents the number of cities in test TSP instances.}
\label{Figure: The generated by MLPR for TSP instances with different size}
\end{figure}
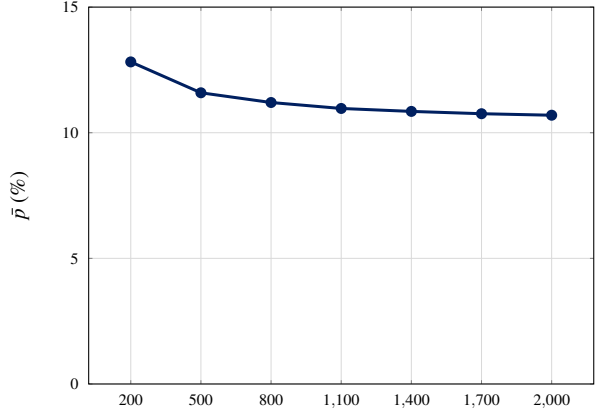

Surprisingly, the optimality gap generated by our MLPR model for each test category is always zero. As the dimension of the test TSP instances increases, our MLPR model tends to prune slightly more edges, as shown in Figure~\ref{Figure: The generated by MLPR for TSP instances with different size}. This suggests that our MLPR model trained on small randomly generated TSP instances generalize well to larger randomly generated TSP instances. Note that the large instances tested here are of similar problem characteristics with the small instances used in training, as they are generated in a similar way. This indicates our MLPR method is likely to work well on practical applications where similar problem instances need to be solved regularly. However, when tested on a problem instance that is different from the training instances, our MLPR method may not  capture the original optimal solution in the reduced problem space, due to its heuristic nature.

\subsection{Varying Problem Types}\label{Subsection: Generalization Regarding Problem Variants}
\subsubsection{Setting}
In this subsection, we train our MLPR model on the symmetric TSP and test it on other TSP variants. We take the first 50 LKCC-hard instances ($\mathcal{I}_6$) from MATILDA as our training set, because overall it generates the smallest optimality gap for symmetric TSP instances with different problem characteristics according to Table \ref{Table: the optimality gap}. We train our MLPR model using SVM with an RBF kernel and test two penalty parameter values $\epsilon_m = 10$ and $\epsilon_m = 100$. We gradually differentiate the test instances away from the training instances, by using 1) symmetric TSP, 2) asymmetric TSP, and 3) sequential ordering problem (SOP) instances in testing. The asymmetric TSP is a TSP variant that allows the distance matrix to be asymmetric; and SOP is a variant that further considers precedence constraints in the order of visiting cities. These test instances are all from the TSP library \citep{reinelt1991tsplib}. For symmetric TSP, we use $19$ instances for which the number of cities is in between $100$ and $200$. For asymmetric TSP and SOP, we use the easy instances that can be solved to optimality by CPLEX with $8$ CPUs in 1000 seconds. We use the trained MLPR model to reduce problem size for each test instance, and solve the original and reduced problems to optimality by exact solvers (Concorde for symmetric TSP, and CPLEX for asymmetric TSP and SOP instances). The MIP formulation used for asymmetric TSP is the Miller-Tucker-Zemlin formulation presented in Section \ref{Subsection: Travelling Salesman Problem}, and the one for SOP is an adaption of the Miller-Tucker-Zemlin formulation with precedence constraints \citep{sherali2002tightening}. The random sampling method used for SOP is  presented in Appendix A. As before, the random sampling and problem reduction process is repeated 25 times to allow for randomness. 



\subsubsection{Results}
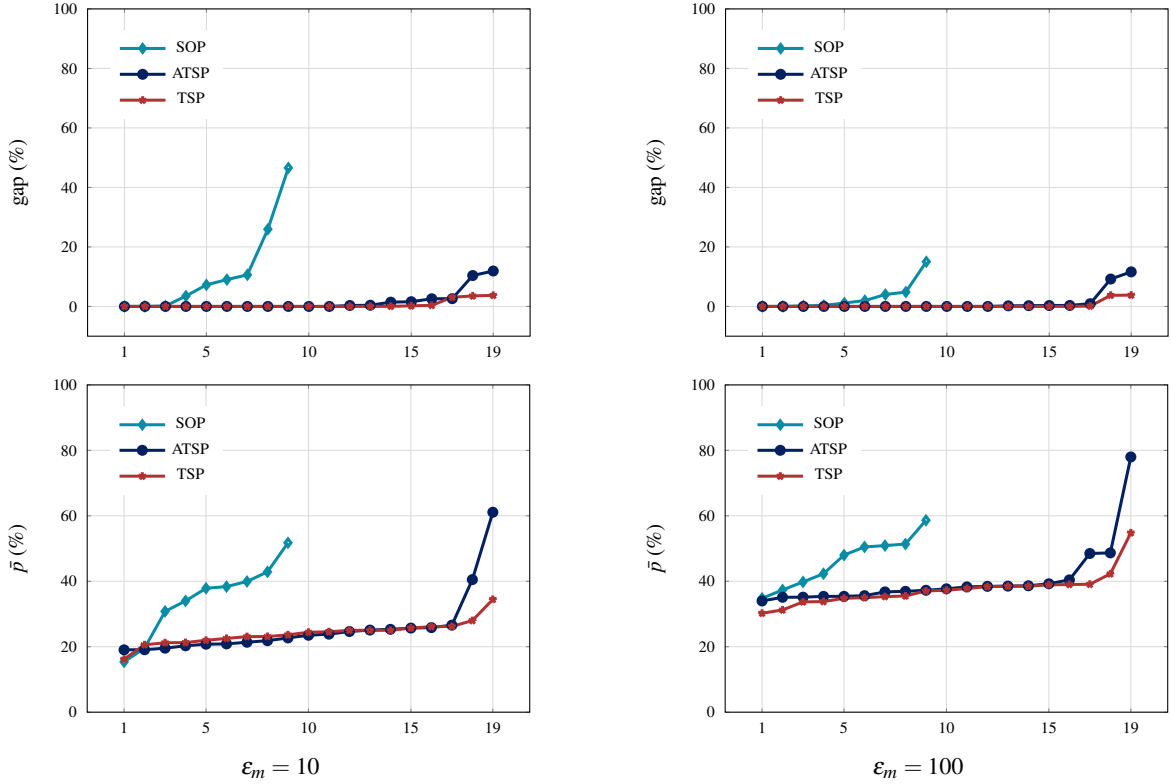
\begin{figure*}[!t]
  \centering
  \captionsetup[subfigure]{labelformat=empty}
  \subfloat[$\epsilon_m=10$]{
    \begin{minipage}[t]{0.45\textwidth}
	\begin{tikzpicture}
	\begin{axis} [box plot width=0.20em, xtick={1,5,10,15,19}, ytick={0,20,40,60,80,100}, ymax = 100, ylabel = \scriptsize gap (\%), height=0.80\textwidth,width=\textwidth, grid style={line width=.1pt, draw=gray!10},major grid style={line width=.2pt,draw=gray!30}, xmajorgrids=true, ymajorgrids=true,  major tick length=0.05cm, minor tick length=0.0cm, legend style={at={(0.05,0.80)},anchor=west,font=\scriptsize,draw=none}]
	\addplot[color=blue,mark=diamond, mark size = 1.5, line width=0.40mm] table{\SOPTenGap};\addlegendentry{\tiny SOP}	
	\addplot[ color=marine,mark=*, mark size = 1.5, line width=0.40mm] table{\ATSPTenGap};\addlegendentry{\tiny ATSP}
	\addplot[color=maroon,mark=star, mark size = 1.5, line width=0.40mm] table{\TSPTenGap};\addlegendentry{\tiny TSP}				
	\end{axis}
      \end{tikzpicture}
      
	\begin{tikzpicture}
	\begin{axis} [box plot width=0.20em, xtick={1,5,10,15,19}, ytick={0,20,40,60,80,100}, ylabel = \scriptsize$\bar{p}$ (\%), ymax = 100,  ymin = 0,  height=0.80\textwidth,width=\textwidth, grid style={line width=.1pt, draw=gray!10},major grid style={line width=.2pt,draw=gray!30}, xmajorgrids=true, ymajorgrids=true,  major tick length=0.05cm, minor tick length=0.0cm, legend style={at={(0.05,0.80)},anchor=west,font=\scriptsize,draw=none}]
	\addplot[color=blue,mark=diamond, mark size = 1.5, line width=0.40mm] table{\SOPTenP};\addlegendentry{\tiny SOP}	
	\addplot[ color=marine,mark=*, mark size = 1.5, line width=0.40mm] table{\ATSPTenP};\addlegendentry{\tiny ATSP}
	\addplot[color=maroon,mark=star, mark size = 1.5, line width=0.40mm] table{\TSPTenP};\addlegendentry{\tiny TSP}						
	\end{axis}
      \end{tikzpicture}
    \end{minipage}
    }
    \hspace{0.7cm}
      \subfloat[$\epsilon_m=100$]{      
        \begin{minipage}[t]{0.45\textwidth}
          \begin{tikzpicture}
	\begin{axis} [box plot width=0.20em, xtick={1,5,10,15,19}, ytick={0,20,40,60,80,100}, ylabel = \scriptsize gap (\%),  ymax = 100, height=0.80\textwidth,width=\textwidth, grid style={line width=.1pt, draw=gray!10},major grid style={line width=.2pt,draw=gray!30}, xmajorgrids=true, ymajorgrids=true,  major tick length=0.05cm, minor tick length=0.0cm, legend style={at={(0.05,0.80)},anchor=west,font=\scriptsize,draw=none}]
	\addplot[color=blue,mark=diamond, mark size = 1.5, line width=0.40mm] table{\SOPHunGap};\addlegendentry{\tiny SOP}	
	\addplot[ color=marine,mark=*, mark size = 1.5, line width=0.40mm] table{\ATSPHunGap};\addlegendentry{\tiny ATSP}
	\addplot[color=maroon,mark=star, mark size = 1.5, line width=0.40mm] table{\TSPHunGap};\addlegendentry{\tiny TSP}				
	\end{axis}
	\end{tikzpicture}\\
	\begin{tikzpicture}
	\begin{axis} [box plot width=0.20em, xtick={1,5,10,15,19}, ytick={0,20,40,60,80,100}, ylabel = \scriptsize $\bar{p}$ (\%), ymax = 100,  ymin = 0, height=0.80\textwidth,width=\textwidth, grid style={line width=.1pt, draw=gray!10},major grid style={line width=.2pt,draw=gray!30}, xmajorgrids=true, ymajorgrids=true,  major tick length=0.05cm, minor tick length=0.0cm, legend style={at={(0.05,0.80)},anchor=west,font=\scriptsize,draw=none}]
	\addplot[color=blue,mark=diamond, mark size = 1.5, line width=0.40mm] table{\SOPHunP};\addlegendentry{\tiny SOP}	
	\addplot[color=marine,mark=*, mark size = 1.5, line width=0.40mm] table{\ATSPHunP};\addlegendentry{\tiny ATSP}
	\addplot[color=maroon,mark=star, mark size = 1.5, line width=0.40mm] table{\TSPHunP};\addlegendentry{\tiny TSP}				
	\end{axis}
	\end{tikzpicture}
      \end{minipage}
    }
    \caption{The optimality gap generated and percentage of remaining problem size after reduction ($\bar{p}$) when training our MLPR model with two values of $\epsilon_m$ on symmetric TSP instances and testing it on symmetric TSP,  asymmetric TSP and SOP instances. The horizontal axis represents the index of test problem instances. We sort the test instances according to the optimality gap (or $\bar{p}$) in ascending order for better visualisation. 
    } 
	\label{Figure: variants instance}
\end{figure*}

\begin{figure*}
	\centering
	\small
	\captionsetup[subfigure]{labelformat=empty}
	\subfloat[\hspace{-1.6cm} average optimiality gap in percentage]{
		\begin{tikzpicture}
		\begin{axis}[ybar, symbolic x coords={gap10, gap100}, xtick=data,  ymax = 15,  ymin = 0, ytick=\empty,  axis y line = none, axis x line* = bottom, nodes near coords={\pgfmathprintnumber[fixed zerofill, precision=2]{\pgfplotspointmeta}}, bar width=24pt, enlarge x limits = 0.5, height=0.45\textwidth,width=0.50\textwidth, xticklabels={\scriptsize $\epsilon_m=10$, \scriptsize $\epsilon_m=100$}, major tick length=0.0cm, minor tick length=0.0cm, legend style={at={(1.05,0.95)},anchor=north,legend columns=0, font=\scriptsize,draw=none}]
		\addplot[color=maroon, fill=maroon] table[x=Category, y=TSP] {\VariantMeanGap};\addlegendentry{\tiny TSP}	
		\addplot[color=marine, fill=marine] table[x=Category, y=ATSP] {\VariantMeanGap};\addlegendentry{\tiny ATSP}	
		\addplot[color=blue, fill=blue] table[x=Category, y=SOP] {\VariantMeanGap};\addlegendentry{\tiny SOP}	
		\end{axis}
		\end{tikzpicture}
	}
	\hspace{-1.6cm}
	\subfloat[average percentage of  remaining problem size]{
		\begin{tikzpicture}
		\begin{axis}[ybar, symbolic x coords={p10, p100},xtick=data,  ymax = 50,  ymin = 0, axis y line = none, axis x line* = bottom,  nodes near coords={\pgfmathprintnumber[fixed zerofill, precision=2]{\pgfplotspointmeta}}, bar width=24pt, enlarge x limits = 0.5, height=0.45\textwidth,width=0.50\textwidth, xticklabels={\scriptsize $\epsilon_m=10$, \scriptsize $\epsilon_m=100$}, major tick length=0.0cm, minor tick length=0.0cm, legend style={at={(0.65,0.40)},anchor=west,font=\scriptsize,draw=none}]
		\addplot[color=maroon, fill=maroon] table[x=Category, y=TSP] {\VariantMeanP};
		\addplot[color=marine, fill=marine] table[x=Category, y=ATSP] {\VariantMeanP};
		\addplot[color=blue, fill=blue] table[x=Category, y=SOP] {\VariantMeanP};
		\end{axis}
		\end{tikzpicture}
	}
	\caption{The average optimality gap generated by our MLPR model and the average percentage of remaining problem size after reduction when training our MLPR model on symmetric TSP instances and testing it on symmetric TSP,  asymmetric TSP and SOP instances.} 
	\label{Figure: average gap and size on problem variants}
\end{figure*}
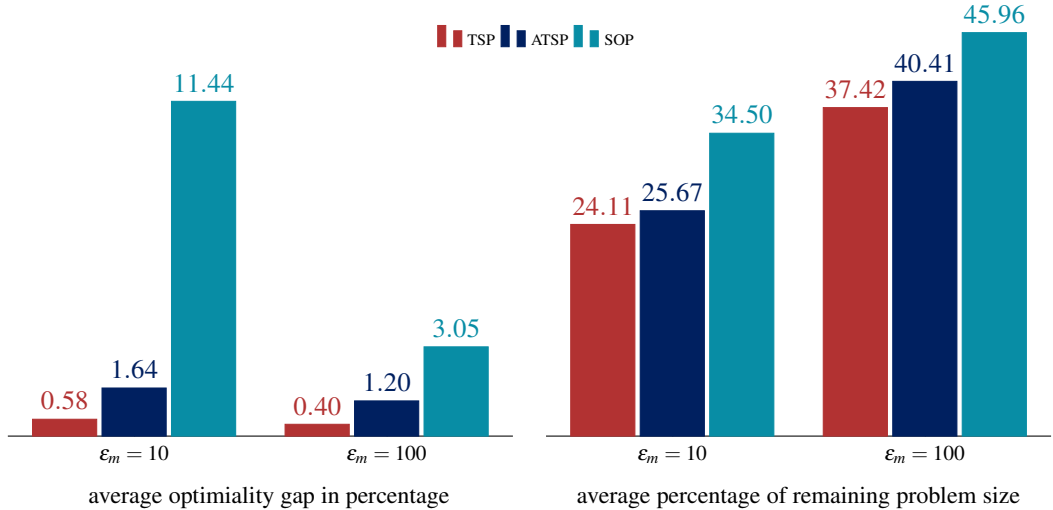

The optimality gap generated by our MLPR model and remaining problem size after reduction for each test instance are presented in Figure \ref{Figure: variants instance}, and the average statistics across each problem type are presented in Figure \ref{Figure: average gap and size on problem variants}. These results show that the MLPR model trained on symmetric TSP instances makes useful predictions about which variables can be eliminated without significantly impacting solution quality when testing on instances from asymmetric TSP and SOP. We also observe that as we gradually move the test instances away from the training instances, our MLPR model becomes less accurate, resulting in a larger optimality gap. On the other hand, it also becomes less confident at pruning edges, resulting in a larger remaining problem size.  When we use a larger penalty $\epsilon_m = 100$, our MLPR model prunes less edges so the optimality gap  generated is smaller.

When the training and test instances are both symmetric TSP's, our MLPR model performs well; it prunes on average 75.89\% of edges but still captures a near-optimal solution in the reduced problem, which is within 0.56\% from the optimal solution. 

When training on symmetric TSP and testing on  asymmetric TSP instances, the optimality gaps generated by our MLPR model (with $\epsilon_m = 100$) are all less than 1\%, except for two instances rbg403 and rbg443. We then use two closely-related asymmetric TSP instances rbg323 and rbg358 as training set, and investigate whether the performance of our MLPR model can be improved for rbg403 and rbg443. The results show that by using asymmetric TSP as training instances, the optimality gap generated by our MLPR model (with $\epsilon_m = 10$) for rbg403 and rbg443 can be significantly reduced from 10.39\% to 1.2\% and from 11.91 \% to 1.25\% respectively.

When training on symmetric TSP and testing on SOP instances, our MLPR model does not perform very well on some instances e.g., ESC12, br17.10 and rbg109a. However we show that by using the other 6 SOP instances as training set, the optimality gaps generated by our MLPR model (with $\epsilon_m=10$) for ESC12, br17.10 and rbg109a can be significantly reduced to 0.65\%, 0.00\%  and 0.23\% respectively.

\subsection{Boosting the Performance of CPLEX}\label{Subsection: Boosting the Performance of CPLEX}
\subsubsection{Setting}
In this subsection, we investigate whether our MLPR model can be used as a preprocessing technique to boost the performance of a generic solver -- CPLEX. Because the symmetric TSP can be efficiently solved by a specialized solver -- Concorde, and an asymmetric TSP can be easily converted to a symmetric TSP~\citep{jonker1983}, we mainly focus on solving SOPs here. 


\begin{table*}[!t]
	\centering
	\caption{The optimization results (in percentage) of CPLEX  and CPLEX-MLPR when used to solve the hard SOP instances within the cutoff time 1000 seconds. The columns gap$_\mathrm{b}$ and gap$_\mathrm{m}$ compute the best and average optimality gaps of the feasible solutions found by CPLEX or CPLEX-MLPR comparing to the lower bounds published in the TSP library; while gap$_\mathrm{s}$ computes an optimality gap based on the lower bounds generated by CPLEX-MLPR; $r$ is the successful rate of finding a feasible solution; $\bar{p}$ is the percentage of remaining problem size after reduction. The last column presents the optimality gap published in the TSP library. We highlight the 4 instances for which the best optimality gap generated by CPLEX-MLPR is better than the gap given in the TSP library.}
	\label{Table: CPLEX results on SOP}
	\begin{tabular}{@{\extracolsep{4pt}}lrrrrrrrr@{}}
		\toprule
		\multirow{2}{*}{Dataset} & \multicolumn{3}{c}{CPLEX} &  \multicolumn{4}{c}{CPLEX-MLPR ($\epsilon_m = 10$)} &  TSPLIB \\\cline{2-4}\cline{5-8}\cline{9-9}
		& gap$_\mathrm{b}$ & gap$_\mathrm{m}$ & $r$\ \  & gap$_\mathrm{b}$ & gap$_\mathrm{m}$ & gap$_\mathrm{s}$ & $\bar{p}\  \ \ $ & gap\ \ \\\midrule
		ft53.1 & 1.25  & 3.52  & 100 & 1.25   & 3.88   & 9.91    & 77.75 & 1.25  \\
		ft53.2 & 8.52  & 17.21 & 68.00  & \bf{6.08}   & 13.18  & 26.83   & 76.77 & 9.24  \\
		ft70.2 & 36.91 & 36.91 & 4.00   & 34.48  & 38.42  & 8.90    & 60.39 & 31.50 \\
		p43.2 &  1.12  & 11.68 & 36.00  & 1.24   & 1.62   & 1532.40 & 79.51 & 0.55  \\
		p43.3 & 1.78  & 64.38 & 12.00  & 1.83   & 40.16  & 2929.60 & 79.36 & 1.11  \\
		p43.4 &  19.31 & 19.38 & 84.00  & 19.31  & 19.69  & 189.35  & 79.70 & 19.25 \\
		rbg150a & 0.11  & 0.18  & 92.00  & 0.11   & 0.85   & 2.46    & 36.94 & 0.11  \\
		ry48p.1 & 3.84  & 5.82  & 100 & 3.84   & 6.14   & 11.51   & 55.48 & 3.84  \\
		ry48p.2 & 8.71  & 11.15 & 100 & 8.03   & 11.51  & 19.33   & 53.84 & 7.36  \\
		ry48p.3 &  13.31 & 27.55 & 12.00  & 12.68  & 21.34  & 45.42   & 47.28 & 9.57  \\
		ry48p.4 & 5.92  & 6.93  & 24.00  & 6.52   & 10.56  & 69.29   & 43.61 & 4.94  \\\midrule
		ft53.3 & -- & --  & 0.00   & \bf{13.62}  & 23.24  & 60.75   & 68.90 & 15.43 \\
		ft70.3 & -- & --  & 0.00   & 8.48   & 15.10  & 18.53   & 57.99 & 2.98  \\
		ft70.4 & -- & --  & 0.00   & 4.83   & 6.98   & 21.03   & 56.39 & 2.47  \\
		kro124p.1 & -- & --  & 0.00   & \bf{5.78}   & 18.74  & 23.73   & 44.10 & 6.53  \\
		kro124p.2 & -- & --  & 0.00   & 21.72  & 43.93  & 51.70   & 44.34 & 8.16  \\
		kro124p.3 & -- & --  & 0.00   & 75.39  & 140.99 & 167.87  & 42.31 & 24.19 \\
		kro124p.4 & -- & --  & 0.00   & 43.40  & 56.23  & 130.46  & 38.69 & 17.34 \\
		prob.100 & -- & --  & 0.00   & 219.82 & 617.78 & 649.68  & 48.00 & 35.25 \\
		rbg253a & -- & --  & 0.00   & \bf{1.91}   & 3.80   & 6.31    & 34.31 & 2.02  \\
		rbg323a & -- & --  & 0.00 & 5.39   & 9.98   & 11.47   & 33.24 & 0.67  \\
		rbg341a & -- & --  & 0.00 & 31.07  & 44.18  & 60.24   & 34.97 & 2.12  \\
		rbg358a & -- & --  & 0.00& 52.18  & 70.17  & 87.45   & 34.77 & 3.22  \\
		rbg378a & -- & -- & 0.00 & 70.66  & 90.92  & 124.25  & 33.18 & 2.61 \\\bottomrule
	\end{tabular}
\end{table*}

We take the $9$ easy SOP instances used in Section \ref{Subsection: Generalization Regarding Problem Variants} as our training set, and train a machine learning model using SVM with RBF kernel ($\epsilon_m = 10$), which takes a few minutes. We then apply the trained model to reduce problem size for hard SOP instances from the TSP library. We select 24 hard SOP instances for testing, which have not been proved to optimality according to the bounds published in the TSP library. The reduction time for the largest test instance (with 378 cities) is around 25 seconds. We use CPLEX with $8$ CPUs to solve the reduced problem, compared to directly solving the original problem. The cutoff time is set to 1000 seconds for both, and the reduction time used by our MLPR model is counted as part of the cutoff time. Note that we also parallelize the process of generating random samples and computing statistical features for our MLPR model.  The MIP emphasis parameter is set to ``HIDDENFEAS" and MIP search method is set to ``TRADITIONAL" for CPLEX, with an emphasis on searching for high-quality feasible solutions. The experiments are again repeated 25 times with different random seeds. We report the successful rate of finding any feasible solution for the 25 independent runs and compute an average optimality gap only for the successful runs. 



\subsubsection{Results}
The optimization results of CPLEX and CPLEX-MLPR when used to solve the SOP instances are presented in Table~\ref{Table: CPLEX results on SOP}. We observe that by using our MLPR method as a preprocessing technique, CPLEX can generally find a comparable or better solution within the cutoff time, especially for hard instances. For $13$ out of $24$ instances, CPLEX fails to find a feasible solution in any of the $25$ runs, i.e., the successful rate is $0$. In contrast, CPLEX-MLPR can always find a feasible solution. This is partially because we feed the best feasible solution from sampling to CPLEX as a warm start.


It is important to note that when using our MLPR method as a preprocessing technique, the generic solver CPLEX can find a better primal solution that improves the best objective value published in the TSP library for $4$ instances (highlighted in bold in Table~\ref{Table: CPLEX results on SOP}). The best objective values found by CPLEX-MLPR for some instances are much larger than the lower bounds published in the TSP library, resulting in a large optimality gap (gap$_\mathrm{b}$ and gap$_\mathrm{m}$). However, this is mainly due to the weak MIP model we used to formulate SOP,  which can be inferred from the fact that the lower bounds produced by CPLEX-MLPR are usually much smaller than the lower bounds published in the TSP library. Thus CPLEX-MLPR can potentially find a better primal solution if using a stronger MIP formulation or simply given more computational time. As our main goal is not to come up with the best CPLEX model for solving SOP, a further investigation along this line is beyond the scope of this paper.

\subsection{Comparing to Other Generic Problem Reduction Methods}
\subsubsection{Setting}
In this subsection, we compare our MLPR method against 
a single correlation-based measure (CBM) described in Section \ref{Subsection: Statistical Measures}, as well as a generic problem reduction method: Construct, Merge, Solve \& Adapt (CMSA) \citep{blum2016construct}. For CBM, we remove edges that are positively correlated with the objective values from a graph. The CMSA method removes edges that do not appear in the sample solutions generated by a probabilistic model, which selects a candidate edge $e_{i,j}$ with probability proportional to $1/(1+c_{i,j})$ when constructing a sampling solution. Hence, edges with a small cost are more likely to be selected in sampling. Note that the CMSA method usually involves multiple iterations. Here, as we are interested in a preprocessing technique, we only compare our MLPR method with the first iteration of CMSA. The sample size for CMSA is set to $n$. We apply our MLPR-$\mathcal{I}_6$ model (trained on LKCC-hard instances), CBM, and CMSA to reduce the problem size for each of the 6 TSP instances used in Table \ref{Table: The ratio between the time taken by CPLEX to solve the original and reduced problem instances.}, and use CPLEX to solve the original and reduced problem instances to optimality. 

\begin{figure}[!t]
	\centering
	\begin{tikzpicture}
	\begin{axis} [box plot width=0.20em,  ylabel = \scriptsize average optimiality gap in percentage (\%), xlabel = \scriptsize average percentage of remaining problem size $\bar{p}$ (\%),  height=0.50\textwidth,width=0.70\textwidth, grid style={line width=.1pt, draw=gray!10},major grid style={line width=.2pt,draw=gray!30}, xmajorgrids=true, ymajorgrids=true,  major tick length=0.05cm, minor tick length=0.0cm, legend style={at={(1.05,0.50)},anchor=west,font=\scriptsize,draw=none}]
	\addplot[ only marks, orange, mark=*, mark size = 2.0] table[ y index = 0, x index = 1]{\TSPCPLEX}; \addlegendentry{MLPR}
	\addplot[only marks, maroon, mark=diamond*, mark size = 2.0] table[ y index = 2, x index = 3]{\TSPCPLEX}; \addlegendentry{CBM}
	\addplot[only marks, marine, mark=triangle*, mark size = 2.0] table[ y index = 4, x index = 5]{\TSPCPLEX}; 	\addlegendentry{CMSA}			
	\end{axis}
	\end{tikzpicture}
	\caption{The remaining problem size versus the optimality gap generated by each problem reduction method, MLPR, CBM, and CMSA on the 6 TSP instances tested.}
	\label{Figure: comparison with baselines}
\end{figure}
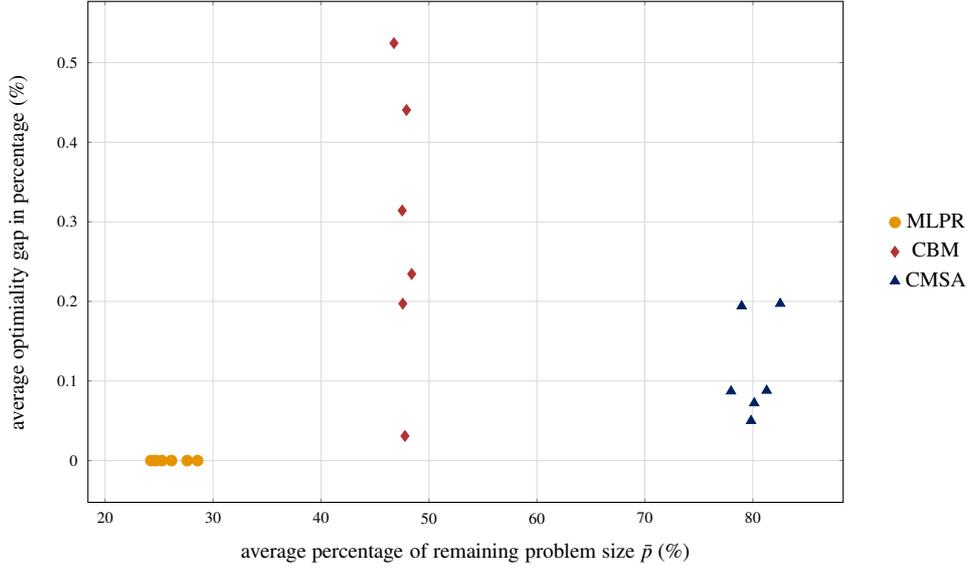

\subsubsection{Results}
The plot of remaining problem size after reduction versus the optimality gap generated by each method is shown in Fig. \ref{Figure: comparison with baselines}.  Our MLPR method outperforms both the CBM and CMSA methods on the 6 instances tested, in the sense that it removes 70\%-80\% of edges and consistently achieves 0\% optimality gap. The CBM and CMSA methods work well in terms of optimality gap (always less than 0.5\%), however, they prune much fewer edges from an instance compared to our MLPR method. Although it is possible to make these two methods more aggressive in pruning edges via parameter tuning (e.g., using a smaller sample size for CMSA), this will further degrade the solution quality generated in the reduced problem instance, resulting in a larger optimality gap.

\section{Conclusion}
In this research, we have applied machine learning techniques to reduce problem size for combinatorial optimization, which can be used as an effective preprocessing step to improve the performance of existing solution algorithms. We have adapted the machine learning model that we originally developed in \citep{sun2019using} to prune the search space for the travelling salesman problem (TSP). We empirically investigated the generalization error of our machine learning model when training and test (unseen) instances have different instance characteristics, sizes or are from different TSP variants. Our experimental results showed that our model generalized reasonably well to a wide range of instances with different characteristics or sizes. In general, when training and test instances are from the same TSP variant, the generalization error of our model is small; and this generalization error naturally increases when testing our model on TSP variants that become increasingly dissimilar to the training instances. Solving a completely different problem variant than in the training set is a fairly extreme case of mismatch between training and application of the model. Hence this shows that our approach is likely to be quite robust to the more typical changes seen in practice when applying a model to, e.g., a business application where the problem characteristics slowly drift over time. In future, we plan to develop a more generic model that does not require re-training when applied to solve a class of problems. Another possible direction for future work would be to apply our problem reduction method to real-world problems where a fast computation of a high-quality solution is desired, such as trip planning. 




\section*{Acknowledgement}
This work was supported by an ARC Discovery Grant (DP180101170) from Australian Research Council.

\section*{Appendix A: Random Sampling Method for Sequential Ordering Problem}

\begin{algorithm}[!t]
	\caption{\textsc{Random Sampling Method} ($V$, $C$, $\mathcal{S}$, $m$)}
	\label{Algorithm: Random Sampling Methods}
	\begin{algorithmic}[1]
	    \Require vertex set $V$; edge cost set $C$; precedence constraint set $\mathcal{S}$; number of samples to generate $m$.
		\State Initialize array $A[i] \leftarrow 0$, $i = 1,2,\cdots |V|$; \Comment{count number of precedences before visiting city $i$.}
		\State Initialize linked lists $L[i] \leftarrow \emptyset$, $i = 1,2,\cdots |V|$; \Comment{denote the cities that should be visited after city $i$.} 
		\For{$k$ from $1$ to $|\mathcal{S}|$} \label{line 1}  
		\State $(i,j) \leftarrow S[k]$; 
		\State Add $j$ to linked list $L[i]$; 
		\State $A[j] \leftarrow A[j] + 1$; 
		\EndFor \label{line 2}  
		\For{$k$ from $1$ to $m$}\Comment{generate $m$ random sample routes.}
		\State Copy the array $A$ to array $A_c$:  $A_c \leftarrow A$; 
		\State Set the initial candidate vertex set $V_c$ to city $1$: $V_c \leftarrow 1$; \Comment{assume routes start from city $1$.}
		\For{$j$ from $1$ to $|V|$}
		\State Randomly select a vertex $v$ from $V_c$;
		\State Add $v$ to the $k_{th}$ sample route $P_k$: $P_k[j] \leftarrow v$; \Comment{$P_k$ denotes the $k_{th}$ sample route.}
		\State Delete $v$ from $V_c$; \Comment{swap $v$ with the last element in $V_c'$ and delete it to gain efficiency.}
		\For{$v'$ in the linked list $L[v]$}\label{line 3}
		\State $A_c[v'] \leftarrow A_c[v'] - 1$; 
		\If{$A_c[v'] == 0$}
		\State Add $v'$ to $V_c$; 
		\EndIf 
		\EndFor\label{line 4}
		\EndFor
		\EndFor \\ 			
		\Return $\{P_1,P_2,\cdots,P_{m}\}$. 
	\end{algorithmic}
\end{algorithm}

The main steps of our random sampling method to generate one feasible route for SOP can be summarized as follows:
\begin{enumerate}
    \item Initialize a route starting from city $1$;  
    \item Compute a set of candidate cities $V_c$ that do not have any precedence after removing the cities that have already been visited;  
    \item Randomly select a city from the candidates $V_c$ to visit; 
    \item Repeat Step 2 and 3 until all cities have been visited. 
\end{enumerate}
To avoid redundant computation, we first iterate through the set of precedence constraints $\mathcal{S}$ to count the number of cities that should be visited before visiting city $i$ ($i=1,\cdots,n$) and store this in array $A$. We also store the individual cities that should be visited after city $i$ ($i=1,\cdots,n$) in a linked list $L$ (line \ref{line 1} to \ref{line 2} in Algorithm \ref{Algorithm: Random Sampling Methods}). Having $A$ and $L$, we can efficiently update the set of candidate cities $V_c$ that can be visited in the next step after removing the cities already visited (line \ref{line 3} to \ref{line 4} in Algorithm \ref{Algorithm: Random Sampling Methods}). The idea is that after removing city $v$ in the current step, we iterate through the linked list $L[v]$ and for every $v'$ in $L[v]$, we decrement $A[v']$ by $1$. If $A[v']$ is equal to $0$, then city $v'$ can be visited in the next step since it does not have any precedence apart from the cities already visited. By doing this, we can generate one sample route in $\mathcal{O}\big(|\mathcal{S}|\big)$ time. Thus the total time complexity of generating $m$ samples is $\mathcal{O}\big(m|\mathcal{S}|\big)$.

\section*{Reference}
\bibliographystyle{apalike}
\bibliography{TSP}

\end{document}